# Deep learning achieves perfect anomaly detection on 108,308 retinal images including unlearned diseases

Ayaka Suzuki & Yoshiro Suzuki*

Tokyo Institute of Technology

**Summary**

Optical coherence tomography (OCT) scanning is useful in detecting various retinal diseases. However, there are not enough ophthalmologists who can diagnose retinal OCT images. To provide OCT screening inexpensively and extensively, an automated diagnosis system is indispensable. Although many machine learning techniques exist, no technique can achieve perfect diagnosis. As long as a technique might overlook a disease, ophthalmologists must double-check even those images that the technique classifies as normal. Here, we show that our deep-learning-based binary classifier (normal or abnormal) achieved a perfect classification on 108,308 retinal OCT images, i.e., true positive rate = 1.000000 and true negative rate = 1.000000; hence, the area under the ROC curve = 1.0000000 (SOTA performance). Although the test set included three types of diseases, two of these were not used for training. Our work has a sufficient possibility of raising automated diagnosis techniques from "assistant for ophthalmologists" to "independent system without ophthalmologists".

**Keywords**

medical image processing, anomaly detection, artificial intelligence, retina, OCT, deep learning, convolutional neural network, metric learning, machine learning

**Introduction**

As of 2016, there are 39 million people who are blind and 246 million people who are partially sighted in the world (WHO, 2016). Optical coherence tomography (OCT) scanning is useful in detecting various retinal diseases at early stages before visual loss occurs. However, in much of the world, there are not enough highly skilled ophthalmologists who can diagnose OCT images. To provide OCT screening inexpensively and extensively throughout the world, an automated diagnosis system for OCT images is indispensable.

To put an automatic medical diagnosis technique into practical use as an independent diagnosis system without relying on ophthalmologists (NOT just as an assistant for ophthalmologists), the following two requirements are very important.

(i) The technique must not overlook any disease, i.e., we must make its true positive rate close to 1.0 without limit.

(ii) The technique must be able to detect unlearned diseases.

If (ii) is not satisfied, we have to prepare medical data of all diseases to train the technique, which is unrealistic and impossible in general.

Recently, deep learning has made remarkable progress, especially among other machine learning algorithms. Many deep-learning-based techniques have been applied in various medical fields. For instance, they have been utilized to detect retinal diseases from fundus images (Thing et al., 2017; Kermany et al., 2018; Burlina et al., 2017), pediatric pneumonia from chest X-rays (Kermany et al., 2018), tuberculosis from chest X-rays (Lakhani et al., 2017; Thing et al., 2018), lung cancer from 3D low-dose chest computed tomography (Courtiol et al., 2019), and skin cancer from skin images (Esteva et al., 2017).

In addition, several deep learning approaches for retinal OCT data have been proposed. Some of them specify the retinal disease type (Kermany et al., 2018; Kuwayama et al., 2019), and some of them construct a segmentation map of the retina (Fauw et al., 2018; Fang et al., 2017; Lee et al., 2017; Lu et al., 2017; Roy et al., 2017). However, none of the conventional approaches (Kermany et al., 2018; Fauw et al., 2018; Rasti et al., 2018a; Schlegl et al., 2019; Haloi et al., 2018) completely satisfy requirements (i) and (ii). Even if an automated technique can specify the retinal disease type and construct a segmentation map of the retina, ophthalmologists must double-check even medical data that the technique classifies as normal, as long as there is a risk of overlooking a disease with the technique.

Unlike many other conventional approaches, our scheme is just a binary classifier (normal or abnormal) for retinal OCT images and does not specify the disease type. Instead, we give priority to enhancing its binary classification accuracy and its detection sensitivity for unlearned retinal diseases. If the performances of these two tasks are sufficiently high, ophthalmologists do not have to double-check OCT images that our scheme classifies as normal, which reduces the number of images that ophthalmologists must diagnose. In particular, in mass screening for the retina, our scheme can drastically reduce the burden of ophthalmologists because many patients would have normal retinas. Ophthalmologists have to diagnose only retinal OCT images that our scheme classifies as abnormal to determine the disease type and degree of progress of the disease.

## Results
**Key performance indicators for clinical application.** To achieve clinical application of an automated medical diagnosis technique, the technique is not allowed to overlook diseases (abnormalities), including unlearned diseases, i.e., the true positive rate (TPR) of the technique should be close to 1.0 without limit.

$$\text{TPR} = \frac{\text{abnormal data correctly classified as abnormal}}{\text{actual abnormal data}}$$

To achieve this, it is necessary to decrease the threshold for detecting abnormalities. However, decreasing the threshold increases not only the TPR but also the false positive rate (FPR).

$$\text{FPR} = \frac{\text{normal data wrongly classified as abnormal}}{\text{actual normal data}}$$

For the FPR, the lower it is, the better. From the above, "FPR when setting TPR = 1.0 (hereinafter referred to as $\text{FPR}_{@\text{TPR}=1.0}$)" is one of the most important key performance indicators for clinical application. $\text{FPR}_{@\text{TPR}=1.0}$ is the ratio between the number of normal data points wrongly categorized as abnormal and the total number of actual normal data points when setting the threshold such that the TPR = 1.0.

For instance, in a case where $\text{FPR}_{@\text{TPR}=1} = 0.3$, the technique wrongly classifies 30% of normal data as abnormal. Thus, an ophthalmologist must double-check this 30% and correct the incorrect classifications. However, if $\text{FPR}_{@\text{TPR}=1} = 0$, the ophthalmologist does not have to double-check this 30%.

We can compute $\text{FPR}_{@\text{TPR}=1}$ from the receiver operating characteristic (ROC) curve describing the relation between the TPR and FPR. In addition, we can compute the area under the ROC curve (AUC) from the ROC curve. In all the experiments in this study, we use both $\text{FPR}_{@\text{TPR}=1}$ and the AUC as performance indicators of the binary classification (normal or abnormal).

**Datasets.** Let us illustrate two datasets (called datasets α and β) of the two-dimensional retinal OCT images used in this study (see "**Datasets**" in the STAR methods for further details). We tested several cases in which we utilized either dataset α or β alone or both of them together.

Dataset α was provided by Kermany et al. (2018) and includes 108,309 horizontal foveal cuts of retinal OCT scans from 4,686 patients. Dataset α consists of 51,140 normal images and 57,169 abnormal images (choroidal neovascularization (CNV): 37,205, drusen: 8,616, diabetic macular edema (DME): 11,348). For the CNV images, only 37,204 images were used to divide them into four equal parts. For the DME images, we increased them from 11,348 to 20,000 by conducting data augmentation (see "**Data augmentation**" in the STAR methods for further details).

Dataset β was provided by Rasti et al. (2018b) and includes 4,060 retinal OCT scans from 148 patients. From the dataset, we utilized 1,604 normal images and 1,328 abnormal images (dry age-related macular degeneration (dry AMD)). Although dataset β also includes 1,128 DME images, we did not use them since the results of the AUC and $\text{FPR}_{@\text{TPR}=1}$ (false positive rate at true positive rate = 1.0) were better compared to when they were used.

For the normal images, we increased them from 1,604 to 60,000 by conducting data augmentation.

In this study, we conducted fourfold cross validation to evaluate the performance of our scheme. To do so, every class in datasets α and β was equally divided into four parts. We thus obtained α1…α4 and β1…β4, as shown in **Fig. 1d**. We used two of them for training, another one for testing, and the fourth one for validating (i.e., tuning) our model. The training, validation, and test sets did not share images. We repeated the evaluation of our scheme's performance under four different combinations of the training, validation, and test sets (**Fig. 1d**).

The validation (tuning) set was used to tune the hyperparameters of our model. For each epoch, we computed the categorical cross-entropy loss of our model using the validation set. We verified the performance of all models saved at each epoch with a validation set and selected the best model to use for testing.

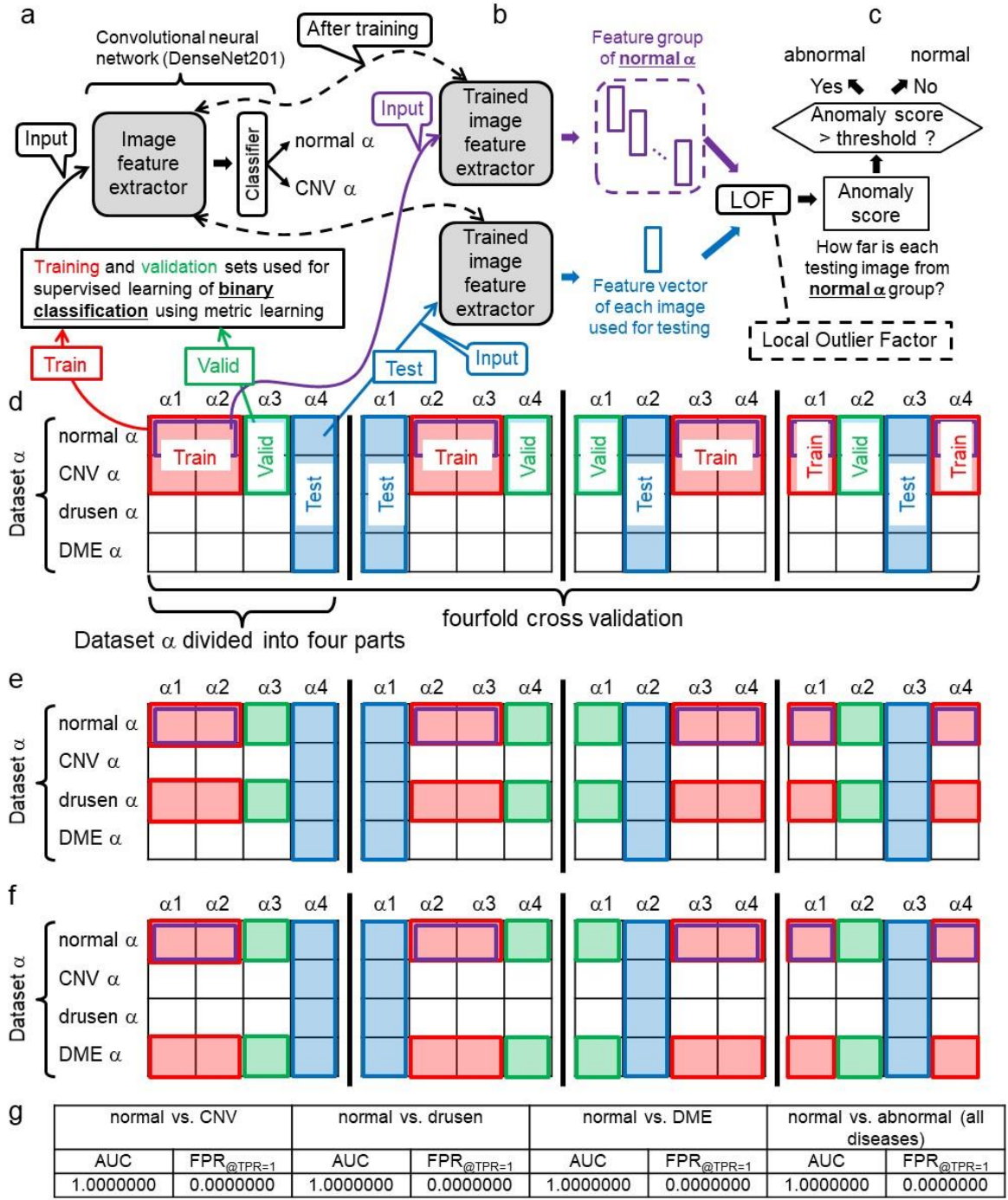

**Fig. 1 | Our AI framework for detecting unlearned diseases. a**, Supervised learning of binary classification with metric learning for our convolutional neural network (CNN) on normal α and CNV α, which represent normal and CNV retinal images in dataset α. **b**, The trained CNN extracts the feature vector from each test image and extracts feature vectors from normal images that are randomly selected from normal α in the training set. **c**, A local outlier factor (LOF) technique computes how far each test image feature is from the feature group of normal α, which is equal to the anomaly score of the test image. When the anomaly score > threshold, the test image is regarded as abnormal. **d**, To evaluate how well our scheme can detect all diseases in dataset α after learning only normal α and CNV α, the procedure illustrated in **Figs. 1a-c** is repeated for the four different combinations of training, validation, and test sets described in **Fig. 1d** (i.e., fourfold cross validation). **e**, Training, validation, and test sets for evaluating how well our scheme can detect all diseases in dataset α after learning only normal α and drusen α. **f**, Training, validation, and test sets for evaluating how well our scheme can detect all diseases in dataset α, after learning only normal α and DME α. **g**, All the results of the AUC (area under the ROC curve) and FPR$_{@TPR=1}$ (false positive rate at true positive rate = 1.0).

**Test set including unlearned diseases.** In this section, we verify whether our scheme can detect unlearned diseases in dataset α. Let us illustrate the algorithm and procedure of our scheme through verification.

By learning *"normality"*, i.e., the characteristic of a normal retina from normal images, our scheme becomes able to detect even unlearned diseases.

In this study, we use a DenseNet201 architecture (Huang et al., 2018) (**Fig. 1a**) as a convolutional neural network (CNN). CNNs are one of the main classes in deep learning and are most commonly applied to analyze images.

First, we prepare a DenseNet201 CNN pretrained on approximately 1.2 million images consisting of 1,000 object categories from the 2012 ImageNet Large Scale Visual Recognition Challenge (Kermany et al., 2018). By retraining the CNN with retinal OCT images, the CNN becomes able to diagnose the retina. In retraining, we update parameters across all layers in the CNN, which is called fine-tuning (Girshick et al., 2014). Kercmany et al. (2018) demonstrate that a CNN pretrained on the ImageNet dataset has a diagnosis performance almost identical to that of a CNN without such pretraining, although the pretrained CNN is retrained on fewer retinal OCT images than another CNN.

In the retraining process shown in **Fig. 1a**, we conduct supervised metric learning of binary classification (normal α or CNV α in the case shown in **Fig. 1d**) for our CNN on normal images and abnormal images with a single type of disease (normal α1, normal α2, CNV α1, and CNV α2 in the case shown in the leftmost figure of **Fig. 1d**). Note that the training set does not include the other two diseases included in dataset α (drusen and DME).

In retraining, we use the Adam optimizer. The loss function is categorical cross entropy. We use neither dropout nor weight decay as the regularization means. The learning rate was 0.001.

The purpose of supervised metric learning (i.e., the retraining) is not to enhance the binary classification accuracy. The true purpose is to gather feature vectors extracted by the CNN from normal images (i.e., feature vectors that are next to the last layer in the CNN outputs) in one place as densely and compactly as possible in the feature vector space. To achieve this purpose, we utilize a metric learning algorithm. We chose the L2-constrained Softmax (Ranjan et al., 2017) from several metric learning algorithms by trial and error (see "**metric learning**" in the STAR methods for further details).

Thus, the CNN learns the "region of normality in the feature space". Note that the training set includes abnormal images as well as normal images (**Fig. 1d**). This enables the CNN to learn a more accurate interface between the normal and abnormal images compared with the case where we train the CNN with only normal images. In other words, use of a training set with abnormal images makes the normal region more compact. Conversely, we subsequently show a case in which the training set includes no abnormal images in **Fig. 4**.

After supervised metric learning, we use the CNN, from which only the last classification layer is removed, as the image feature extractor (**Fig. 1b**). We select 5,000 normal images randomly from the training set and input each of them into the CNN one by one. Consequently, we obtain 5,000 normal feature vectors, which are called the "feature group of normal α", as depicted in **Fig. 1b**.

Next, we input each image in the test set, which consists of normal α4, CNV α4, drusen α4, and DME α4 in the case shown in the leftmost figure of **Fig. 1d**, into the CNN one by one. Finally, a local outlier factor (LOF) (Breunig et al., 2000) technique computes how far the feature of each test image is from the feature group of normal α, which is equal to the anomaly score of the test image. When the anomaly score > threshold, the test image is regarded as abnormal. Therefore, our scheme has the possibility of detecting unlearned diseases that are not included in the training set.

The following datasets are a summary of the above.
- Training set: Half of normal α (e.g., normal α1 and normal α2) and half of images with a single type of disease in dataset α (e.g., CNV α1 and α2) in the case shown in the leftmost figure of **Fig. 1d**.
- Validation set: As depicted in the green frame in the leftmost figure of **Fig. 1d**, another quarter of normal α (e.g., normal α3) and another quarter of CNV α (e.g., CNV α3) that are not used for training.
- Test set: As described in the blue frame in the leftmost figure of **Fig. 1d**, the remaining quarters in dataset α (e.g., normal α4, CNV α4, drusen α4, and DME α4) that are used for neither training nor validation.

To evaluate the performance of our scheme, we conducted the following three fourfold cross validations. First, to evaluate the performance of our scheme trained on normal α and CNV α, we conducted fourfold cross validation, i.e., we repeated the procedure illustrated in **Figs. 1a-c** for the four different combinations of training, validation, and test sets depicted in **Fig. 1d**. Next, we conducted another fourfold cross validation to evaluate the performance of our scheme trained on normal α and drusen α (**Fig. 1e**). Finally, we conducted another fourfold cross validation to evaluate the performance of our scheme trained on normal α and DME α (**Fig. 1f**).

As shown in **Fig. 1g**, all three fourfold cross validations show that our scheme achieved a perfect binary classification (normal or abnormal) on 108,308 retinal OCT images, i.e., true positive rate = 1.00000 and true negative rate = 1.00000; hence, the AUC = 1.0000000 and $FPR_{@TPR=1}$ (false positive rate at true positive rate = 1.0) = 0.00000.

Although the test set includes three types of diseases, two of these are not used for training; nevertheless, all the test images, including unlearned diseases, were correctly classified. In other words,
· When our scheme learns only normal and CNV, it can detect CNV, DME, and drusen.
· When our scheme learns only normal and drusen, it can detect CNV, DME, and drusen.
· When our scheme learns only normal and DME, it

can detect CNV, DME, and drusen.

**Test set including unlearned race.** There is a large difference in the race of patients between datasets α and β (see "**Datasets**" in the STAR methods for further details). Therefore, to investigate whether our scheme can cope with the difference in race, we conducted the following experiment.

As shown in **Fig. 2a**, three-label classification (normal α, normal β and AMD β) is trained by supervised metric learning for our CNN. **Figures 2b and 2c** describe the procedures of extracting image features and testing, which are similar to those shown in **Figs. 1b and 1c**, respectively.

Note that the training, validation, and test sets in **Fig. 2d** differ from those in **Fig. 1d**.
- Training set: Half of normal α (e.g., normal α1 and normal α2) and half of dataset β (e.g., normal β1, normal β2, AMD β1 and AMD β2), as shown in the red frame in the leftmost figure of **Fig. 2d**.
- Validation set: As depicted in the green frame in the leftmost figure of **Fig. 2d**, another quarter of normal α (e.g., normal α3) and another quarter of dataset β (e.g., normal β3 and AMD β3) that are not used for training.
- Test set: As described in the blue frame in the leftmost figure of **Fig. 2d**, the remaining quarters of dataset α (e.g., normal α4, CNV α4, drusen α4, and DME α4) that are not used for training and validation.

As described in **Fig. 2e**, the fourfold cross validation test shows that our scheme achieved an almost perfect binary classification (normal or abnormal) on 108,308 OCT images (AUC = 0.9999841, $FPR_{@TPR=1}$ = 0.0133164).

In other words, even if our scheme learns no abnormal image in dataset α, it can detect all the abnormal images in dataset α with very high accuracy as long as it learns the abnormal images in dataset β, which is independent from dataset α.

Next, we examined whether our scheme can diagnose dataset α without learning dataset α at all, as shown in **Fig. 3**. In other words, we tested a case where normal α was eliminated from the training set in **Fig. 2d**. As described in **Fig. 3e**, our scheme showed considerably low performance in terms of both the AUC and $FPR_{@TPR=1}$ (AUC = 0.4807988, $FPR_{@TPR=1}$ = 1.0000000). Therefore, we demonstrate that to achieve an almost perfect anomaly detection for dataset α, our scheme has to learn at least the normal images in dataset α as well as those in dataset β. This is a natural consequence and is not a fatal defect of our scheme.

**AI trained with only normal images.** As shown in **Fig. 4**, we examined whether our scheme can diagnose dataset α without learning any abnormal images. In other words, we tested a case where we eliminated all the abnormal images from the training set in **Fig. 2d**.

As depicted in **Fig. 4e**, our scheme showed relatively low performance in terms of both the AUC and $FPR_{@TPR=1}$ (AUC = 0.9994540, $FPR_{@TPR=1}$ = 0.1069026) compared with each of **Figs. 1e and 2e**. Therefore, we demonstrate that to achieve perfect anomaly detection for dataset α, our scheme has to learn at least one type of disease.

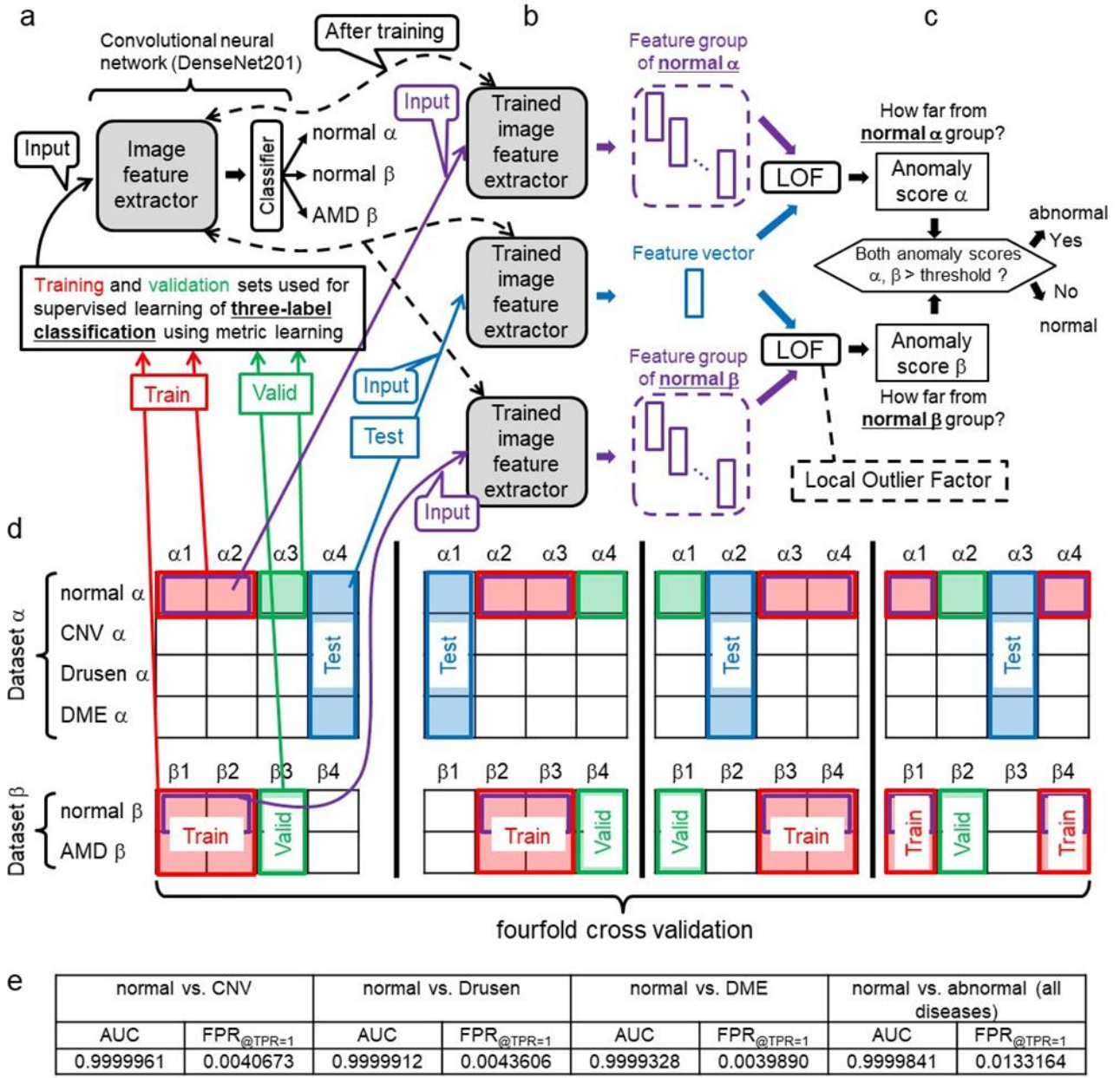

**Fig. 2 | Our AI framework for detecting diseases of unlearned human race. a**, Supervised learning of three-label classification with metric learning for our convolutional neural network (CNN) on normal α, normal β and AMD β. Note that the CNN learns no abnormal image in dataset α. There is a large difference in race of patients between dataset α and β. **b**, The trained CNN extracts feature vectors respectively from test images, normal images that are randomly selected from normal α in the training set and other normal images that are randomly selected from normal β in the training set. **c**, A local outlier factor (LOF) technique computes how far each test image feature is from the feature group of normal α, which is equal to the anomaly score of the test image. The LOF technique also computes how far each test image feature is from the feature group of normal β, which is equal to the anomaly score β. When the both anomaly scores α and β > threshold, the test image is regarded as abnormal. **d**, The procedure illustrated in **Figs. 2a-c** is repeated for the four different combinations of training, validation, and test sets described in **Fig. 2d** (i.e., fourfold cross validation). **e**, All the results of the AUC and $FPR_{@TPR=1}$ (false positive rate at true positive rate = 1.0).

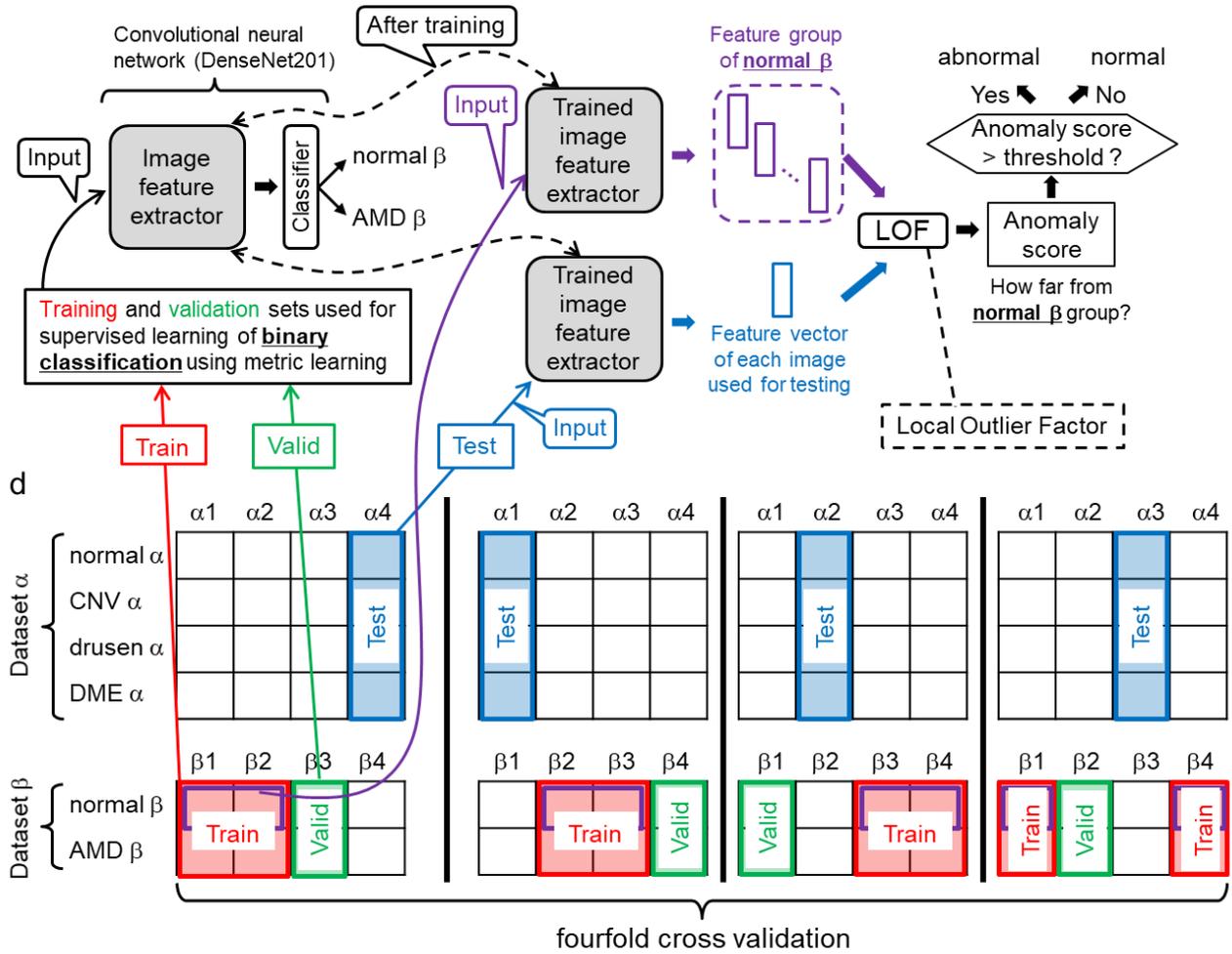

**Fig. 3 | Our AI framework for diagnosing Dataset α after learning only Dataset β. a**, Supervised learning of two-label classification with metric learning for our convolutional neural network (CNN) on dataset β (normal β and AMD β). Note that the CNN learns no image in dataset α. There is a large difference in race of patients between dataset α and β. **b**, The trained CNN extracts the feature vectors from test images and normal images that are randomly selected from normal β in the training set. **c**, A local outlier factor (LOF) technique computes how far each test image feature is from the feature group of normal β, which is equal to the anomaly score of the test image. When the anomaly scores > threshold, the test image is regarded as abnormal. **d**, The procedure illustrated in **Figs. 3a-c** is repeated for the four different combinations of training, validation, and test sets described in **Fig. 3d** (i.e., fourfold cross validation). **e**, All the results of the AUC and $FPR_{@TPR=1}$ (false positive rate at true positive rate = 1.0).

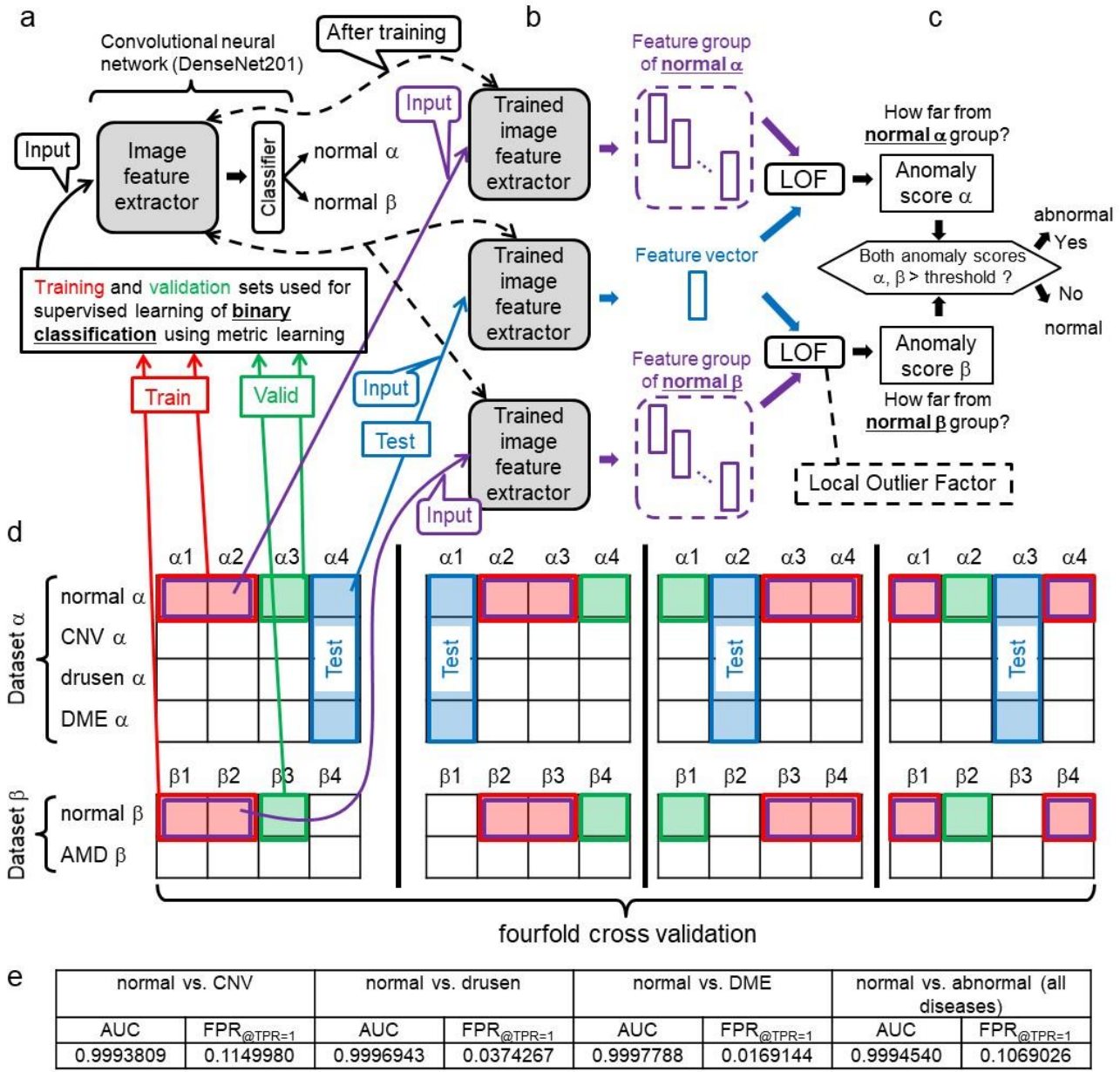

**Fig. 4 | Our AI framework trained on only normal images. a**, Supervised learning of binary classification with metric learning for our convolutional neural network (CNN) on normal α and normal β. **b**, The trained CNN extracts the feature vectors respectively from normal images that are randomly selected from normal α in the training set, and other normal images that are randomly selected from normal β in the training set. **c**, A local outlier factor (LOF) technique computes how far each test image feature is from the feature group of normal α, which is equal to the anomaly score α of the test image. The LOF technique also computes how far each test image feature is from the feature group of normal β, which is equal to the anomaly score β of the test image. When both anomaly scores α and β > threshold, the test image is regarded as abnormal. **d**, The procedure illustrated in **Figs. 4a-c** is repeated for the four different combinations of training, validation, and test sets described in **Fig. 4d** (i.e., fourfold cross validation). **e**, All the results of the AUC and FPR$_{@TPR=1}$ (false positive rate at true positive rate = 1.0).

## Discussion

As shown in **Fig. 1g**, our deep-learning-based binary classifier (normal or abnormal) achieved a perfect classification (AUC = 1.0000000 and $FPR_{@TPR=1}$ = 0.0000000) on 108,308 two-dimensional retinal OCT images when the following requirement was satisfied.

1. The training set includes normal images and images with at least one type of disease obtained from the same hospital as that where the test set is obtained. This indicates that our scheme has a sufficient possibility of detecting unlearned diseases without overlooking abnormalities, which are the most important points for the clinical application of an automated diagnosing system, as illustrated in the introduction.

Conversely, from **Figs. 2e, 3e and 4e**, our scheme did not achieve a perfect classification when either of the following conditions was satisfied.

2. Training set includes both "normal images obtained from the same hospital as that where the test set was obtained" and "normal and abnormal images from another hospital".
3. Training set includes no images obtained from the hospital where the test set was obtained.
4. Training set includes no abnormal images.

The above three points indicate that our scheme cannot perfectly diagnose retinal OCT images obtained all over the world when we train our scheme with only those retinal OCT datasets available via the internet (e.g., datasets α and β). This is a natural consequence because there are differences in retinal structure and retinal lesions between different races of people.

As described in **Supplementary Table 1**, compared with the conventional approaches for diagnosing retinal OCT scans, the anomaly detection performance of our scheme is overwhelmingly high and reliable.

Although there are some limitations related to our scheme and we must conduct further verifications (illustrated in "**Limitations of our scheme**" in the supplementary information), we demonstrate that our scheme has a sufficient possibility of pushing up the automated diagnosis technique for retinal OCT scans from "assistant for ophthalmologists" to "independent diagnosis system without ophthalmologists", which none of the conventional approaches has ever achieved.

The main contribution of this study is that we prove the existence of an automated diagnosis scheme achieving an AUC of 1.0000000 on 108,308 retinal images, including unlearned diseases.

## Acknowledgement

The authors received no specific funding for this work.

All the datasets used for this work, which are illustrated in STAR Methods, are gratefully acknowledged.

## Author contributions

A.S. and Y.S. designed the proposed anomaly detection method.

A.S. contributed to the software engineering and developed most of the network architectures.

Y.S. managed the project, analyzed the results, and wrote the paper.

## Figure Legends

**Fig. 1 | Our AI framework for detecting unlearned diseases. a**, Supervised learning of binary classification with metric learning for our convolutional neural network (CNN) on normal α and CNV α, which represent normal and CNV retinal images in dataset α. **b**, The trained CNN extracts the feature vector from each test image and extracts feature vectors from normal images that are randomly selected from normal α in the training set. **c**, A local outlier factor (LOF) technique computes how far each test image feature is from the feature group of normal α, which is equal to the anomaly score of the test image. When the anomaly score > threshold, the test image is regarded as abnormal. **d**, To evaluate how well our scheme can detect all diseases in dataset α after learning only normal α and CNV α, the procedure illustrated in **Figs. 1a-c** is repeated for the four different combinations of training, validation, and test sets described in **Fig. 1d** (i.e., fourfold cross validation). **e**, Training, validation, and test sets for evaluating how well our scheme can detect all diseases in dataset α after learning only normal α and drusen α. **f**, Training, validation, and test sets for evaluating how well our scheme can detect all diseases in dataset α, after learning only normal α and DME α. **g**, All the results of the AUC (area under the ROC curve) and $FPR_{@TPR=1}$ (false positive rate at true positive rate = 1.0).

**Fig. 2 | Our AI framework for detecting diseases of unlearned human race. a**, Supervised learning of three-label classification with metric learning for our convolutional neural network (CNN) on normal α, normal β and AMD β. Note that the CNN learns no abnormal image in dataset α. There is a large difference in race of patients between dataset α and β. **b**, The trained CNN extracts feature vectors respectively from test images, normal images that are randomly selected from normal α in the training set and other normal images that are randomly selected from normal β in the training set. **c**, A local outlier factor (LOF) technique computes how far each test image feature is from the feature group of normal α, which is equal to the anomaly score of the test image.

The LOF technique also computes how far each test image feature is from the feature group of normal β, which is equal to the anomaly score β. When the both anomaly scores α and β > threshold, the test image is regarded as abnormal. **d**, The procedure illustrated in **Figs. 2a-c** is repeated for the four different combinations of training, validation, and test sets described in **Fig. 2d** (i.e., fourfold cross validation). **e**, All the results of the AUC and $FPR_{@TPR=1}$ (false positive rate at true positive rate = 1.0).

**Fig. 3 | Our AI framework for diagnosing Dataset α after learning only Dataset β. a**, Supervised learning of two-label classification with metric learning for our convolutional neural network (CNN) on dataset β (normal β and AMD β). Note that the CNN learns no image in dataset α. There is a large difference in race of patients between dataset α and β. **b**, The trained CNN extracts the feature vectors from test images and normal images that are randomly selected from normal β in the training set. **c**, A local outlier factor (LOF) technique computes how far each test image feature is from the feature group of normal β, which is equal to the anomaly score of the test image. When the anomaly scores > threshold, the test image is regarded as abnormal. **d**, The procedure illustrated in **Figs. 3a-c** is repeated for the four different combinations of training, validation, and test sets described in **Fig. 3d** (i.e., fourfold cross validation). **e**, All the results of the AUC and $FPR_{@TPR=1}$ (false positive rate at true positive rate = 1.0).

**Fig. 4 | Our AI framework trained on only normal images. a**, Supervised learning of binary classification with metric learning for our convolutional neural network (CNN) on normal α and normal β. **b**, The trained CNN extracts the feature vectors respectively from normal images that are randomly selected from normal α in the training set, and other normal images that are randomly selected from normal β in the training set. **c**, A local outlier factor (LOF) technique computes how far each test image feature is from the feature group of normal α, which is equal to the anomaly score α of the test image. The LOF technique also computes how far each test image feature is from the feature group of normal β, which is equal to the anomaly score β of the test image. When both anomaly scores α and β > threshold, the test image is regarded as abnormal. **d**, The procedure illustrated in **Figs. 4a-c** is repeated for the four different combinations of training, validation, and test sets described in **Fig. 4d** (i.e., fourfold cross validation). **e**, All the results of the AUC and $FPR_{@TPR=1}$ (false positive rate at true positive rate = 1.0).

**STAR Methods**
1. **RESOURCE AVAILABILITY**

**1.1. Lead Contact.**
Further information and requests for resources and reagents should be directed to and will be fulfilled by the Lead Contact, Yoshiro Suzuki (ysuzuki@ginza.mes.titech.ac.jp).

**1.2. Materials Availability.**
Although there are no restrictions for use of the materials disclosed, please be sure to cite this manuscript when you use our methods (even if you use a part of our methods).

**1.3. Data and Code Availability.**
**Dataset.** Let us illustrate the two datasets (datasets α and β) of two-dimensional retinal OCT images used in this study. There is a large difference in the race of patients between datasets α and β. Both datasets are available via the internet. We conduct no image preprocessing, such as denoising, segmentation, and retinal alignment, for datasets α and β.

**Dataset α**
- Resource: Kermany et al. (2018)
- Download:
  https://data.mendeley.com/datasets/rscbjbr9sj/3
- License: Creative Commons Attribution 4.0 International License (CC BY-NC-SA 4.0)
- Country where retinal OCT images were obtained:
  United States (Shiley Eye Institute of the University of California San Diego, the California Retinal Research Foundation, Medical Center Ophthalmology Associates)
  China (Shanghai First People's Hospital and Beijing Tongren Eye Center)
- Exclusion criteria: Data obtained from adult patients only, with no other criteria (gender and race)
- OCT scanning device: Spectralis OCT, Heidelberg Engineering, Germany
- Number of retinal OCT images:
  108,309 horizontal foveal cuts of retinal OCT scans from 4,686 patients
  - Normal: 51,140
  - choroidal neovascularization (CNV): 37,205
    Note that we used only 37,204 images to equally divide the images into four equal parts.
  - drusen: 8,616
  - diabetic macular edema (DME): 11,348
    Note that we increased the DME images from 11,348 to 20,000.

**Dataset β**
- Resource: Rasti et al. (2018b)
- Download:
  https://sites.google.com/site/hosseinrabbanikhorasgani/datasets-1
- License: Nothing is mentioned in the paper. The above download site says only the following: "Please reference the paper if you would like to use any part of datasets and this method."
- Country where retinal OCT images were obtained:
  Iran (Noor Eye Hospital in Tehran)
- OCT scanning device: Heidelberg SD-OCT imaging systems
- Exclusion criteria: Nothing is mentioned in the paper
- Number of retinal OCT images:
  4,142 retinal OCT scans from 148 patients.

X

- Normal: 1,678 (from 50 patients)
  Note that we used only 1,604 images apart from 50 eye-ground images, 23 low-quality images and 1 image that could not be divided into four equal parts. By conducting the data augmentation as illustrated below, the 1,604 images increased to 60,000.
- Dry age-related macular degeneration (dry AMD): 1,524 (from 48 patients)
  Note that we used only 1,328 images apart from 48 eye-ground images, 4 low-quality images, 142 images with the ".PNG" extension and 2 images that could not be divided into four equal parts.
- Diabetic macular edema (DME): 1,186 (from 50 patients)
  Note that we did not use those images at all since the results of the AUC and $FPR_{@TPR=1}$ (false positive rate at true positive rate = 1.0) were better compared to when we used those images.

**Data augmentation.**
We utilize Augmentor (Bloice et al., 2016), a Python package designed for data augmentation, to increase the number of retinal OCT images in datasets α and β. Augmentor uses a pipeline-based approach, where operations are conducted sequentially. Each image in dataset β is passed through the following pipeline, where each operation is applied to the image with a given probability.
  1. Flip each image horizontally with a probability of 0.8.
  2. Rotate each image with a probability of 0.7. The rotation angle ranges from -10 to +10 degrees.

**Code availability.** The code shown here corresponds to Supplementary Fig. 1 in our paper. Although there are no restrictions for use of the code, please be sure to cite this manuscript when you use our code (even if you use a part of our code).

- TensorFlow: https://www.tensorflow.org
- ImageNet: http://www.image-net.org/
- Keras: https://keras.io/
- DenseNet201: https://keras.io/applications/#densenet
- Our Code: https://github.com/SAyaka0122/Deep-learning-based-binary-classifier

## 2. EXPERIMENTAL MODEL AND SUBJECT DETAIS

**Use of human subjects.** All the human subjects used in this study were downloaded via the internet. The license of the human subjects is illustrated in "Datasets" of the Data and Code Availability. In addition, this study was approved by the Tokyo Institute of Technology. The approval number is 2019058. We used human subjects.

## 3. METHOD DETAILS

Our scheme is a combination of the following conventional schemes.

**Learning deep features for one-class (DOC) classification (Perera et al., 2018).** This paper presents a deep-learning-based binary classifier (normal or abnormal) that is trained with labeled images from an unrelated task (e.g., ImageNet and CIFAR-10) as well as normal images from a target task.

First, the authors trained a CNN by supervised learning of multilabel classification including the target normal label and multiple other labels in the unrelated dataset. The purpose of supervised learning is to enable the CNN to extract image features useful for distinguishing normal from abnormal. For this purpose, features extracted from normal images should maintain a low intraclass variance in the feature space.

Next, the authors selected normal images randomly from the training set and input each of them into the trained CNN one by one. Consequently, they obtained a "feature group of normal images". They used a binary classification method such as the one-class SVM, SVDD or k-nearest neighbor to compute how far each test image feature is from the feature group of normal images, which is equal to the anomaly score of the test image. When the anomaly score > threshold, the test image is regarded as abnormal.

This method and our scheme have something in common but the following differences.
- Although the training set in the paper (Perera et al., 2018) includes images that are completely unrelated to the target task, the training set in our scheme does not.
- Although the training set in their paper does not include abnormal images from the target task, the training set in our scheme does.
- Although the authors do not employ a metric learning algorithm to train the CNN, we do in this study.
- Although the authors do not search the optimum layer in the CNN that outputs the image feature vector used for detecting anomalies, we do in this study (see the last paragraph in "**Tuning of our scheme**" in the STAR methods).

**Face reidentification.** Many conventional approaches (Kermany et al., 2018; Kuwayama et al., 2019) specify the disease type based on medical images. These approaches classify diseases both in learning and testing. In general, they cannot classify unlearned diseases in testing.

Conversely, the CNN in our scheme learns to classify retinal diseases in training, but it does not classify the retinal disease in testing. In testing, we use the CNN as just an image feature extractor (see **Fig. 1b**). The trained CNN extracts features from each test image. We compare the feature with other features extracted from normal images and regard the test image as normal if they are similar (**Fig. 1c**). Therefore, our scheme has the possibility of detecting unlearned diseases. Note that our scheme cannot specify the disease type. Our scheme is similar to individual identification techniques called face reidentification (Li et al., 2012).

**Metric learning.** In the face reidentification technique illustrated in the previous section, feature vectors extracted from images of the same person by a CNN should be similar (the distance between the vectors should be short), but the distance between feature vectors extracted from two different persons should be long.

To achieve this, a metric learning technique is often employed when the CNN performs face classification. Metric learning is the task of learning a distance function over objects. If we can conduct the metric learning successfully, feature vectors of the same person are gathered in one place compactly in the feature vector space even if the photographic conditions (angle, light source, facial expression, etc.) are different. In addition, feature vectors extracted from two different persons are different even if the photographic conditions are similar.

When detecting anomalies for retinal OCT images, we must train the CNN so that the distribution of feature vectors from normal images is as compact as possible in the feature vector space.

Until now, special loss functions such as contrastive loss (Rao et al., 2016) and triplet loss (Wang et al., 2014) have been used in mainstream metric learning. However, since 2017, new metric learning methods such as L2-constrained Softmax (Ranjan et al., 2017), ArcFace (Deng et al., 2018), SphereFace (Liu et al., 2017), and CosFace (Wang et al., 2018) have been proposed, and their effectiveness has been demonstrated. These new methods do not use the special loss function but partially change the CNN structure.

We tested the metric learning methods below and found that L2-constrained Softmax is the most suitable for our scheme.
- L2-constrained Softmax
- ArcFace

**Tuning of our scheme.** Our scheme is simple but has high anomaly detection performance. To achieve a high performance, we adjusted the following points of our scheme by trial and error, which are not illustrated in the main text.
· Selecting an optimum convolutional neural network (CNN) architecture (**Figs. 1a, 2a, 3a, 4a and Supplementary Fig. 1a**). We tested the below CNNs and found that DenseNet201 (Huang et al., 2018) is the most suitable for our scheme.
  - DenseNet121, DenseNet169, DenseNet201
  - ResNet50
  - InceptionV3, InceptionResNetV2
  - Xception
  - VGG16
  - MobileNet
· Selecting an optimum metric learning algorithm (**Figs. 1a, 2a, 3a, 4a and Supplementary Fig. 1a**) for supervised learning of our CNN. We tested the metric learning algorithms below and found that L2-constrained Softmax (Ranjan et al., 2017) is the most suitable for our scheme.
  - L2-constrained Softmax
  - ArcFace (Deng et al., 2018)
· Selecting an optimum data augmentation technique for dataset β (**Figs. 2d, 3d, 4d**). We tested the augmentation techniques below and found that the combination of rotation and horizontal flip is suitable for our scheme (see "**Data augmentation**" in the STAR methods for further details).
  - Rotation
  - Horizontal flip
  - Mix-up (Zhang et al., 2017)
  - Test-time augmentation (Simonyan et al., 2014)
· Selecting an optimum layer in our CNN (i.e., DenseNet201 (Huang et al., 2018)) that outputs the image feature vector used for detecting retinal anomalies (**Figs. 1b, 2b, 3b, 4b and Supplementary Fig. 1b**). We tested the feature vectors outputted from the following layers and found that the next to the last layer is the most suitable for our scheme.
  - next to the last layer
  - 10th batch normalization layer from the last layer
  - 45th batch normalization layer from the last layer
  - 80th batch normalization layer from the last layer
  - 115th batch normalization layer from the last layer

Note that all the above layers are contained in the fourth Dense block of DenseNet201. DenseNet201 is composed of 709 layers in total.

## 4. QUANTIFICATION AND STATISTICAL ANALYSIS

Here, we illustrate how to obtain (quantificate) the results of Fig. 1. How to obtain the other results (Figs. 2-4) is almost similar and therefore excluded here.

To achieve clinical application of an automated medical diagnosis technique, the technique is not allowed to overlook diseases (abnormalities), including unlearned diseases, i.e., the true positive rate (TPR) of the technique should be close to 1.0 without limit.

$$\text{TPR} = \frac{\text{abnormal data correctly classified as abnormal}}{\text{actual abnormal data}}$$

To achieve this, it is necessary to decrease the threshold for detecting abnormalities. However, decreasing the threshold increases not only the TPR but also the false positive rate (FPR).

$$\text{FPR} = \frac{\text{normal data wrongly classified as abnormal}}{\text{actual normal data}}$$

For the FPR, the lower it is, the better. From the above, "FPR when setting TPR = 1.0 (hereinafter referred to as $\text{FPR}_{@TPR=1.0}$)" is one of the most important key performance indicators for clinical application. $FPR_{@TPR=1.0}$ is the ratio between the number of normal data points wrongly categorized as abnormal and the total number of actual normal data points when setting the threshold such that the TPR = 1.0.

The 108,308 horizontal foveal cuts of retinal OCT scans from 4,686 patients were used for training, validating, and testing our AI.

We conducted fourfold cross validation to evaluate the performance of our AI. In other words, the training, validation, and test sets did not share images. In each fold of the cross validation, the test set included 27,077 images.

As shown in **Fig. 1g**, all three fourfold cross validations show that our scheme achieved a perfect binary classification (normal or abnormal) on 108,308 retinal OCT images, i.e., true positive rate = 1.00000 and true negative rate = 1.00000; hence, the AUC (area under the ROC curve) = 1.0000000 and $FPR_{@TPR=1}$ (false positive rate at true positive rate = 1.0) = 0.00000.

Although the test set includes three types of diseases, two of these are not used for training; nevertheless, all the test images, including unlearned diseases, were correctly classified. In other words,
- When our scheme learns only normal and CNV, it can detect CNV, DME, and drusen.
- When our scheme learns only normal and drusen, it can detect CNV, DME, and drusen.
- When our scheme learns only normal and DME, it can detect CNV, DME, and drusen.

**Supplementary information**

**Comparison of anomaly detection performances with conventional automated diagnoses.** As shown in **Supplementary Table 1**, a comparison between our scheme and the conventional diagnosis techniques for retinal OCT scans indicates the following points.
- Compared with the conventional approaches, our anomaly detection performance can be regarded as overwhelmingly high and reliable because it achieved perfect anomaly detection (AUC = 1.0000000 and $FPR_{@TPR=1}$ = 0.0000000) on a large test set that includes 108,308 retinal OCT images in total.
- In terms of the AUC, Kermany et al. (2018) achieved 0.999, which is the best among all the conventional methods and almost the same as ours (AUC = 1.0000000). However, $FPR_{@TPR=1}$ of their method is 0.49, which is much higher (i.e., worse) than ours ($FPR_{@TPR=1}$ = 0.0000000). In other words, when setting the anomaly detection threshold such that their method overlooks no disease at all, their method classifies 49% of normal data as abnormal. In addition, their method does not have a function to detect unlearned diseases, and their method was tested on a small dataset consisting of only 1,000 retinal OCT images.
- In terms of $FPR_{@TPR=1}$, Fauw et al. (2018) achieved 0.38, which is the lowest (i.e., best) among all the conventional methods but much higher than ours ($FPR_{@TPR=1}$ = 0.0000000). In addition, their method does not have a function to detect unlearned diseases, and their method was tested on a small dataset consisting of only 997 data points. In general, $FPR_{@TPR=1}$ has a tendency to increase when the test set becomes large. This is because a larger test set has a higher possibility of including "special and rare data that appear normal but are actually abnormal". To correctly classify this rare data as abnormal, we have to decrease the threshold for detecting abnormalities, which would increase $FPR_{@TPR=1}$.
- Although we tested our scheme using 108,308 retinal OCT images, all conventional algorithms except for f-AnoGAN (Schlegl et al., 2019) were tested on much smaller datasets of 1,000 or fewer images. The authors tested f-AnoGAN on 70,000 partial retinal OCT images, but its performance was much lower (AUC = 0.93 and $FPR_{@TPR=1}$ > 0.8) than that of ours.

**Supplementary Table 1 | Comparison of anomaly detection performances for retinal OCT images**

| Reference | AUC (area under the ROC curve) | $FPR_{@TPR=1}$ (false positive rate at true positive rate = 1.0) | Classification task | Number of images in the test set |
|---|---|---|---|---|
| Ours (**Fig. 1e**) | 1.0000000 | 0.0000000 | normal vs. abnormal (test set includes unlearned diseases) | 108,308 |
| Ours (**Fig. 2e**) | 0.9999841 | 0.0133164 | normal vs. abnormal (test set includes abnormal images of unlearned race) | 108,308 |
| Ours (**Supplementary Fig. 1**) | 1.00000 | 0.00000 | normal vs. abnormal (test and training sets are independent but test set has no unlearned disease) | 1,000 |
| Kermany et al., 2018 | 0.999 | 0.49 | urgent referral vs. the others (test and training sets are independent, but test set has no unlearned disease) | 1,000 |
| Fauw et al., 2018 | 0.9921 | 0.38 | urgent referral vs. the others (test and training sets are independent, but test set has no unlearned disease) | 997 |
| Rasti et al., 2018a | 0.989-0.993 | Unknown | normal vs. abnormal | 45 - 60 |
| Schlegl et al., 2019 | 0.9301 | > 0.8 | normal vs. abnormal | 70,000 |
| Haloi et al., 2018 | Unknown | 0.53 | normal vs. abnormal | 1,000 |

**Anomaly detection performance of our scheme on an independent dataset.** As shown in **Figs. 1-4**, we conduct fourfold cross validation to test our scheme. No test set includes images used for training. However, in the cross validation depicted in **Figs. 1, 2, and 4**, we cannot say that the training and test sets are completely independent of each other. In other words, we cannot deny the possibility that the training and test sets share images of the same patient, which are not exactly the same image (e.g., their acquisition periods were different).

Here, we test our scheme in a case where the training and test sets are completely independent, i.e., they do not share images from the same patient.

As shown in **Supplementary Fig. 1d**, we trained our deep learning model (CNN) on all images in dataset α and tested the trained model with another dataset provided by Kermany et al. (2018). This test set is composed of 1,000 retinal OCT images (normal: 250, DME: 250, CNV: 250, drusen: 250) obtained from 633 patients who are independent from the patients of dataset α. As shown in **Supplementary Fig. 1e**, our scheme achieved a perfect binary classification (normal or abnormal) for all 1,000 images, i.e., AUC = 1.00000 and $FPR_{@TPR=1}$ = 0.00000. It is demonstrated that our scheme can cope with a case where the training and test sets are independent.

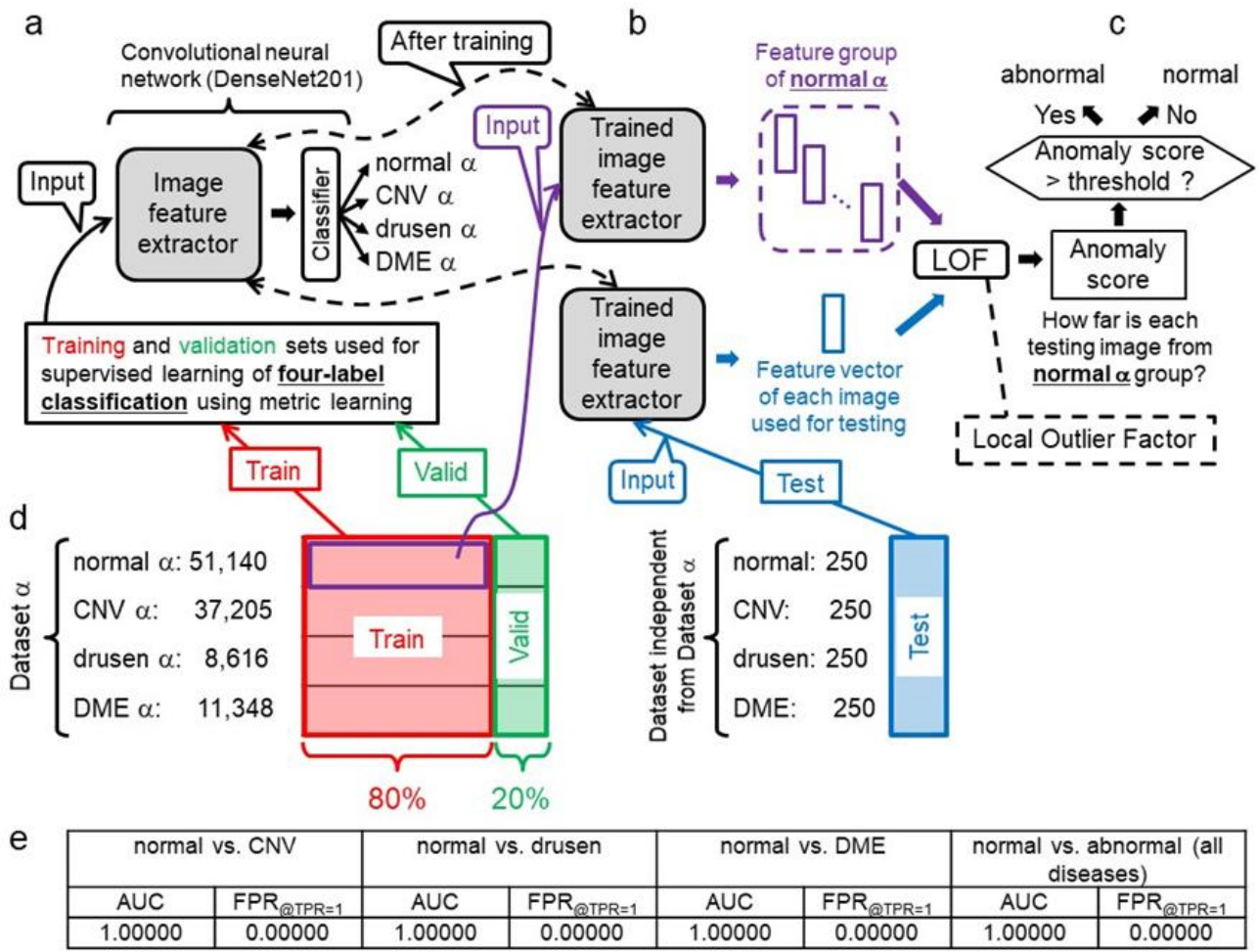

**Supplementary Fig. 1 | Our AI framework trained on an independent dataset. a**, Supervised learning of four-label classification with metric learning for our convolutional neural network (CNN) on dataset α. **b**, The trained CNN extracts the feature vectors from normal images that are randomly selected from normal α in the training set. **c**, The test set is independent from dataset α. A local outlier factor (LOF) technique computes how far each test image feature is from the feature group of normal α, which is equal to the anomaly score of the test image. When the anomaly score > threshold, the test image is regarded as abnormal. **d**, Training, validation, and test sets for evaluating how well our scheme can detect all diseases in the independent dataset. **e**, All the results of the AUC (area under the ROC curve) and $FPR_{@TPR=1}$ (false positive rate at true positive rate = 1.0).

**Limitations of our scheme.** The limitations of our scheme are as follows.

- Our scheme classifies a retinal OCT image as either normal or abnormal only. It cannot determine the disease type nor the degree of progress of the disease. The function of our scheme is only to reduce the number of OCT images that an ophthalmologist must double-check. Ophthalmologists do not have to double-check OCT images that our scheme classifies as normal because our scheme does not overlook anomalies. If we give our scheme the extra ability to specify the retinal disease type, the possibility that our scheme fails to detect unlearned diseases would increase. In addition, even if our scheme could specify disease types, including unlearned diseases, ophthalmologists would eventually have to double-check all OCT images that our scheme classifies as abnormal to evaluate the degree of progress of the disease and determine medical treatment.
- In all the experiments in this study, we examined only whether our scheme can detect choroidal neovascularization (CNV), diabetic macular edema (DME), and drusen. In future work, we will investigate whether our scheme can correctly detect other important retinal diseases.
- In all the experiments in this study, both the training and test sets are retinal images obtained by OCT scanning devices provided by the same company, i.e., Heidelberg Engineering, Germany. In future work, we will investigate whether our scheme can correctly diagnose images obtained by another company's OCT device after our scheme is trained with images obtained by an OCT device provided by Heidelberg Engineering.
- In all the experiments in this study, we utilized the retinal OCT images provided by Kermany et al. (2018) (i.e., dataset α) as the test set. The reasons why other datasets were not used for testing are as follows.
    - The image labeling for dataset α is highly reliable. Six or more ophthalmologists (including two senior specialists with over 20 years of clinical retina experience) conducted a three-stage labeling procedure (Kermany et al., 2018).
    - Dataset α consists of more than 100,000 retinal OCT images.
    - Dataset α is available via the internet.

    From the above, dataset α is the most reliable and largest retinal OCT dataset available via the internet. As shown by the results in **Supplementary Table 1**, our scheme yielded no misjudgment for all 108,308 images in dataset α, which indicates that our scheme is also highly reliable. However, if we utilize different unreliable datasets for testing, the following problem might occur. When our scheme cannot achieve perfect anomaly detection for an unreliable test set, we will not be able to distinguish our scheme's misjudgment from the incorrect labeling of an ophthalmologist (i.e., misjudgment of ophthalmologist who constructed the dataset). Therefore, we utilized only dataset α as the test set in all the experiments in this paper. However, if we can obtain another dataset as reliable as dataset α, we will investigate whether our scheme can correctly diagnose the dataset.